\documentclass{article}

\usepackage{graphicx}

\usepackage[nonatbib,final]{neurips_2022}

\usepackage{wrapfig}
\usepackage[utf8]{inputenc} 
\usepackage[T1]{fontenc}    
\usepackage{hyperref}       
\usepackage{url}  
\usepackage{booktabs}       
\usepackage{amsfonts}       
\usepackage{nicefrac}       
\usepackage{microtype}      
\usepackage{xcolor} 
\usepackage{amsmath}
\usepackage{amsthm}
\newtheorem{theorem}{Theorem}

\usepackage{wrapfig}
\title{Heatmap Distribution Matching for Human Pose Estimation}

\author{%
  Haoxuan Qu \\
  SUTD \\
  Singapore \\
  \texttt{haoxuan\_qu@mymail.sutd.edu.sg} \\
  \And
  Li Xu \\
  SUTD \\
  Singapore \\
  \texttt{li\_xu@mymail.sutd.edu.sg} \\
  \And
  Yujun Cai \\
  NTU \\
  Singapore \\
  \texttt{yujun001@e.ntu.edu.sg} \\
  \And
  Lin Geng Foo \\
  SUTD \\
  Singapore \\
  \texttt{lingeng\_foo@mymail.sutd.edu.sg} \\
  \And
  Jun Liu \thanks{Corresponding Author}\\
  SUTD \\
  Singapore \\
  \texttt{jun\_liu@sutd.edu.sg} \\
}

\begin{document}

\maketitle

\begin{abstract}
For tackling the task of 2D human pose estimation, the great majority of the recent methods regard this task as a heatmap estimation problem, and optimize the heatmap prediction using the Gaussian-smoothed heatmap as the optimization objective and using the pixel-wise loss (e.g. MSE) as the loss function. In this paper, we show that optimizing the heatmap prediction in such a way, the model performance of body joint localization, which is the intrinsic objective of this task, may not be consistently improved during the optimization process of the heatmap prediction. To address this problem, from a novel perspective, we propose to formulate the optimization of the heatmap prediction as a distribution matching problem between the predicted heatmap and the dot annotation of the body joint directly. By doing so, our proposed method does not need to construct the Gaussian-smoothed heatmap and can achieve a more consistent model performance improvement during the optimization of the heatmap prediction. We show the effectiveness of our proposed method through extensive experiments on the COCO dataset and the MPII dataset.

\end{abstract}

\section{Introduction}

2D human pose estimation aims to locate body joints of a person in a given RGB image. It is relevant to a variety of applications, such as action recognition \cite{yan2018spatial}, human-machine interaction \cite{zhang2014pose}, and sign language understanding \cite{moryossef2020real}. For tackling the task of 2D human pose estimation, most of the recent methods \cite{tompson2014joint,newell2016stacked,xiao2018simple,sun2019deep,li2019rethinking,cheng2020higherhrnet,zhang2020distribution,yang2020transpose,li2021tokenpose,luo2021rethinking,YuanFHLZCW21} are \textit{heatmap-based}, i.e., they regard 2D human pose estimation as a heatmap estimation problem. Specifically, for each body joint, these methods generally estimate a grid-like heatmap, on which each pixel value represents the probability that this pixel contains the body joint. Compared to the methods \cite{toshev2014deeppose,carreira2016human,wei2020point,li2021human} that directly regress the coordinates of body joints (i.e. \textit{coordinate regression-based methods}), the \textit{heatmap-based methods} demonstrate a more robust performance since they maintain the spatial structure of the input image throughout the encoding and decoding process \cite{gu2021removing}. 

\begin{figure}[t]
\centering
\includegraphics[width=0.9\textwidth]{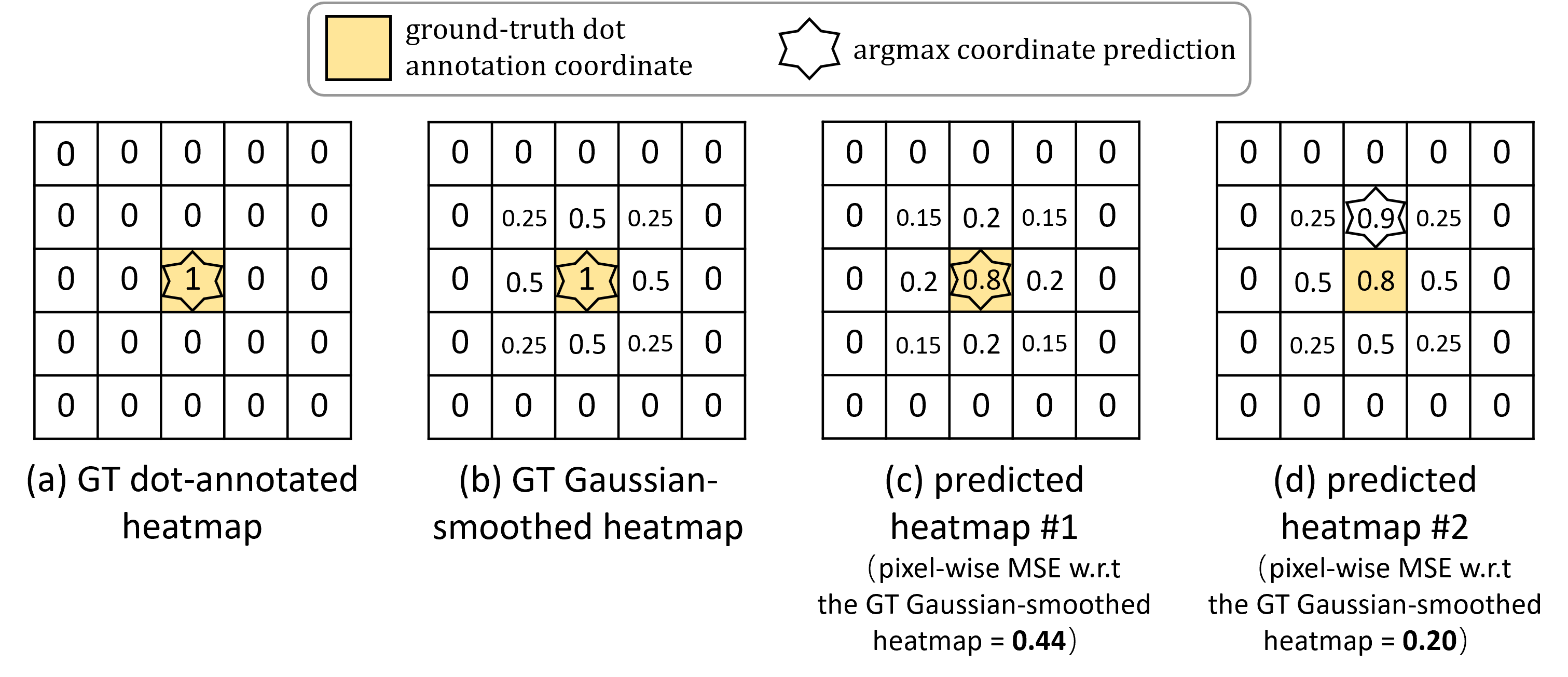}
\caption{Illustration of heatmaps. Although the pixel-wise MSE loss calculated between the predicted heatmap \#2 and the Gaussian-smoothed heatmap is smaller than the loss calculated between the predicted heatmap \#1 and the Gaussian-smoothed heatmap, the predicted heatmap \#2 localizes the body joint wrongly, whereas the predicted heatmap \#1 localizes the body joint correctly.
}
\label{fig:example}
\end{figure}

During the training process of the \textit{heatmap-based methods}, an important step is the optimization of the heatmap prediction. This optimization can be done naively via constructing a dot-annotated heatmap for each body joint as shown in Fig.~\ref{fig:example}(a), and then measuring the difference (i.e., conducting pixel-wise comparison) between the predicted heatmap and the constructed ground-truth (GT) dot-annotated heatmap. However, such a dot-annotated heatmap is sparse, as it has the same zero value for all pixels except the pixel representing the dot annotation of the body joint. Because of this, optimizing the heatmap prediction in such a naive way can lead to a hard training process and a suboptimal model performance \cite{szegedy2016rethinking}. To address this problem, most \textit{heatmap-based methods} \cite{tompson2014joint,newell2016stacked,xiao2018simple,sun2019deep,li2019rethinking,cheng2020higherhrnet,zhang2020distribution,yang2020transpose,li2021tokenpose,luo2021rethinking,YuanFHLZCW21} adopt a strategy to construct Gaussian-smoothed heatmaps, where pixels near the dot annotation have larger pixel values than pixels far from the dot annotation. Specifically, they construct the Gaussian-smoothed heatmap via smoothing the dot annotation of the body joint through a Gaussian distribution as shown in Fig.~\ref{fig:example}(b), instead of only setting the pixel representing the dot annotation to be one. 

While easing the training process, constructing the GT Gaussian-smoothed heatmap still brings problems into the model training process. 
Firstly, for constructing the Gaussian-smoothed heatmap, we need to choose a proper standard deviation of the Gaussian distributions. However, the proper standard deviations of the Gaussian distributions (i.e., the standard deviations that can lead to an optimal performance) often vary across different types of body joints, different body postures, and different body sizes \cite{luo2021rethinking}. Hence, the standard deviations of the Gaussian distributions often need to be carefully chosen, which is non-trivial. 
Secondly, during the process of optimizing the heatmap prediction by minimizing the pixel-wise loss (e.g. MSE) between the predicted heatmap and the Gaussian-smoothed heatmap, the model performance of body joint localization may not be consistently improved. As shown in Fig.~\ref{fig:example}, although compared to the loss calculated between the predicted heatmap \#1 and the Gaussian-smoothed heatmap, the pixel-wise MSE loss calculated between the predicted heatmap \#2 and the Gaussian-smoothed heatmap is smaller, the predicted heatmap \#2 localizes the body joint wrongly, whereas the predicted heatmap \#1 localizes the body joint correctly.

As a result, optimizing the heatmap prediction using the dot-annotated heatmap and the Gaussian-smoothed heatmap as the ground-truth both have their respective problems. Hence, in this work, we aim to tackle their respective problems, and propose to optimize the heatmap prediction directly via minimizing the difference between the predicted heatmap and the dot annotation. By doing so, we can optimize the model directly towards 
accurately localizing the dot annotation 
of the body joint, which is the intrinsic objective of 2D human pose estimation, instead of optimizing the model indirectly towards either the dot-annotated heatmap or the Gaussian-smoothed heatmap. However, as the number of pixels in the predicted heatmap and the number of entries representing the dot annotation are different, we cannot measure the difference between the predicted heatmap and the dot annotation trivially by measuring their entry-wise difference. To handle this problem, inspired by the fact that we can measure the difference between two distributions via measuring their Earth Mover's Distance even if they have different numbers of entries, in this paper, we propose to first formulate the optimization of the heatmap prediction as a distribution matching problem. Specifically, we construct two distributions respectively from the predicted heatmap and the dot annotation. After that, we optimize the heatmap prediction via minimizing the distribution difference 
based on the Earth Mover's Distance. Using such a novel method to optimize the heatmap prediction directly from the dot annotation, we do not need to construct the Gaussian-smoothed heatmap, as well as avoiding the issues of the binary dot-annotated heatmap. Thus, our method achieves superior performance.

Our proposed method is simple yet effective, which can be easily applied to various off-the-shelf 2D human pose estimation models by replacing their original loss function with our proposed loss function measuring the distribution difference between the predicted heatmap and the dot annotation. We experiment our proposed method on multiple models and our method achieves a consistent model performance improvement. 

The contributions of our work are summarized as follows.
1) 
We analyze (in Sec.~\ref{Sec:proof}) 
that the performance of the human pose estimation model may not be consistently improved during the process of minimizing the pixel-wise loss between the predicted heatmap and the GT Gaussian-smoothed heatmap.
2)
From a novel perspective, we formulate the optimization of the heatmap prediction as a distribution matching problem between the predicted heatmap and the GT dot annotation directly, which bypasses the step of constructing the Gaussian-smoothed heatmap and achieves consistent model performance improvement.
3)
Our proposed method achieves state-of-the-art performance on the evaluated benchmarks.

\section{Related Work}

\noindent\textbf{2D Human Pose Estimation.} Due to the wide range of applications, the task of 2D human pose estimation has received lots of attention \cite{toshev2014deeppose,carreira2016human,wei2020point,li2021human,nibali2018numerical,tompson2014joint,newell2016stacked,xiao2018simple,sun2019deep,li2019rethinking,cheng2020higherhrnet,zhang2020distribution,yang2020transpose,li2021tokenpose,luo2021rethinking,YuanFHLZCW21,sun2018integral,gu2021removing,gu2022dive}. 
DeepPose \cite{toshev2014deeppose} made the first attempt of applying deep neural networks into the task of 2D human pose estimation via directly regressing the coordinates of body joints. This type of \textit{coordinate regression-based methods} \cite{toshev2014deeppose,carreira2016human,wei2020point,li2021human} often show inferior performances compared to the \textit{heatmap-based methods} \cite{tompson2014joint,newell2016stacked,xiao2018simple,sun2019deep,li2019rethinking,cheng2020higherhrnet,zhang2020distribution,yang2020transpose,li2021tokenpose,luo2021rethinking,YuanFHLZCW21}, as the \textit{heatmap-based methods} can preserve the spatial structure of the input image throughout the encoding and decoding process \cite{gu2021removing}. Hence, recently, the great majority of the state-of-the-art methods \cite{tompson2014joint,newell2016stacked,xiao2018simple,sun2019deep,li2019rethinking,cheng2020higherhrnet,zhang2020distribution,yang2020transpose,li2021tokenpose,luo2021rethinking,YuanFHLZCW21} regard 2D human pose estimation as a heatmap estimation problem instead of the coordinate regression problem. Among the \textit{heatmap-based methods}, Tompson et al. \cite{tompson2014joint} proposed to apply Markov Random Field (MRF) into the task of 2D human pose estimation. After that, an "hourglass" network, with a conv-deconv architecture, was proposed by Newell et al. \cite{newell2016stacked}. Xiao et al. \cite{xiao2018simple} proposed a baseline method to predict the heatmap via adding several deconvolutional layers to a backbone network. Later on, to maintain high-resolution representations throughout the heatmap estimation process, HRNet was proposed by Sun et al. \cite{sun2019deep}. Yuan et al. \cite{YuanFHLZCW21} further proposed HRFormer to learn the high-resolution representations utilizing a transformer-based architecture. 
Besides the above \textit{heatmap-based methods} that use the Gaussian-smoothed heatmap as the optimization objective, there are also some methods \cite{sun2018integral,gu2021removing,gu2022dive} that combine the idea of \textit{heatmap} and \textit{coordinate regression} by taking the expectation of the predicted heatmap as the predicted coordinates.

Here in this work, different from previous works, our method bypasses both the step of regression and the step of constructing the Gaussian-smoothed heatmap as the optimization objective. Instead, from a novel perspective, we propose to formulate the optimization of the heatmap prediction as a distribution matching problem by minimizing the distribution difference between the predicted heatmap and the dot annotation.

\noindent\textbf{Distribution Matching.} The idea of distribution matching has been studied in various tasks \cite{rubner2000earth,schulter2017deep,wang2020distribution,zhang2020deepemd,zhou2020occlusion,peng2022optimal}, such as image retrieval \cite{rubner2000earth}, tracking \cite{schulter2017deep}, few-shot learning \cite{zhang2020deepemd}, and long-tail recognition \cite{peng2022optimal}. In this work, from a novel perspective, we design a new distribution matching scheme to optimize the heatmap prediction with the help of sub-pixel resolutions for 2D human pose estimation.

\section{Method}

In 2D human pose estimation, optimizing the heatmap prediction using the dot-annotated heatmap and the Gaussian-smoothed heatmap as the ground-truth both have their respective problems. Hence, in this work, we aim to handle their respective problems, and optimize the heatmap prediction directly with the dot annotation of the body joint. To achieve this goal, we propose to formulate the optimization of the heatmap prediction as a distribution matching problem, and minimize the difference between the distribution constructed from the predicted heatmap and that constructed from the dot annotation based on the Earth Mover's Distance.

Below, we first give a brief review of the Earth Mover’s Distance, and then discuss how we formulate the heatmap optimization process as a distribution matching problem. After that, we introduce how we construct the loss function measuring the distribution difference.

\subsection{Revisiting Earth Mover's Distance}
\label{Sec:pre}

The Earth Mover's Distance is a 
a technique used for measuring the difference between two probability distributions, which can be understood as the optimal cost needed to transport the mass from one distribution to another. Specifically, for calculating the Earth Mover's Distance, we regard the source distribution as a set of ($N$) suppliers $S = (s_1, ..., s_N)^\top$, where $s_n$ represents the total units of mass that the $n$-th supplier has, and we regard the target distribution as a set of ($M$) demanders $D = (d_1, ..., d_M)^\top$, where $d_m$ represents the total units of mass that the $m$-th demander requires. Besides, we also denote $C \in  R^{N \times M}_{\geq 0}$ as the cost function between the source and target distributions, where $C_{n,m}$ represents the cost for transporting a unit of mass from the $n$-th supplier to the $m$-th demander. Then we aim to find a least-cost transportation plan from the set of possible plans $P = \{p \in R^{N \times M}_{\geq 0}:~p\textbf{1} = S~\&~p^\top\textbf{1} = D\}$ to transport all mass from the $N$ suppliers to the $M$ demanders, where \textbf{1} represents a vector of ones. The Earth Mover's Distance $E_C(S, D)$ denotes the cost of the least-cost transportation plan, which can be formulated as:
\begin{equation}\label{eq:pre_1}
\begin{aligned}
E_C(S, D)~=~\min_{p \in P}\quad \langle C,p \rangle
\end{aligned}
\end{equation}
where $\langle \cdot , \cdot \rangle$ represents the Frobenius dot product.

However, optimizing Eq.~\ref{eq:pre_1} directly is computationally expensive. Hence, to reduce the computational cost especially when handling large-scale problems, Cuturi \cite{cuturi2013sinkhorn} proposed a regularized formulation of the Earth Mover's Distance $E_C^{reg}(S, D)$ as:
\begin{equation}\label{eq:pre_2}
\begin{aligned}
E_C^{reg}(S, D)~=~\langle C,p^{reg} \rangle
~\textbf{where}~ p^{reg}~=~\mathop{\arg\min}_{p \in P}\quad  \left[\langle C,p \rangle - \frac{1}{\lambda} h(p)\right]
\end{aligned}
\end{equation}
where $\lambda > 0$, and $h(p) = -\sum_{n=1}^{N} \sum_{m=1}^{M} C_{n,m} \log C_{n,m}$. Optimizing Eq.~\ref{eq:pre_2} is computationally cheaper than optimizing Eq.~\ref{eq:pre_1}, since Eq.~\ref{eq:pre_2} can be optimized with matrix scaling through the Sinkhorn algorithm.

\subsection{Problem Formulation}
\label{Sec:form}
Below, we formulate the optimization of the heatmap prediction as a distribution matching problem by constructing the suppliers $S$, the demanders $D$, and the cost function $C$. 

\noindent\textbf{The suppliers $S$.} For a body joint, let $H_{pred} \in R^{H \times W}$ denote its corresponding predicted heatmap with height $H$ and width $W$. To formulate the suppliers $S$ to represent $H_{pred}$, we first localize a set of $H \times W$ suppliers corresponding to the
$H \times W$ pixels of $H_{pred}$. 
Then, for determining the units of mass each supplier stores, as the suppliers cannot hold negative units of mass, we construct $S$ based on a non-nagetive formulation of $H_{pred}$. Besides, we also constrain the total units of mass stored by the suppliers $S$ to be the same as the total units of mass required by the demanders $D$. To meet these requirements, we derive $S$ by first passing $H_{pred}$ through a $relu$ activation function and then normalizing it as:
\begin{equation}\label{eq:supplier}
\begin{aligned}
S~=~\frac{relu(H_{pred})}{\left\|relu(H_{pred})\right\|_1}
\end{aligned}
\end{equation}

\begin{wrapfigure}[20]{r}{0.5\textwidth}
\vspace{-0.4cm}
\centering
\includegraphics[width=0.5\textwidth]{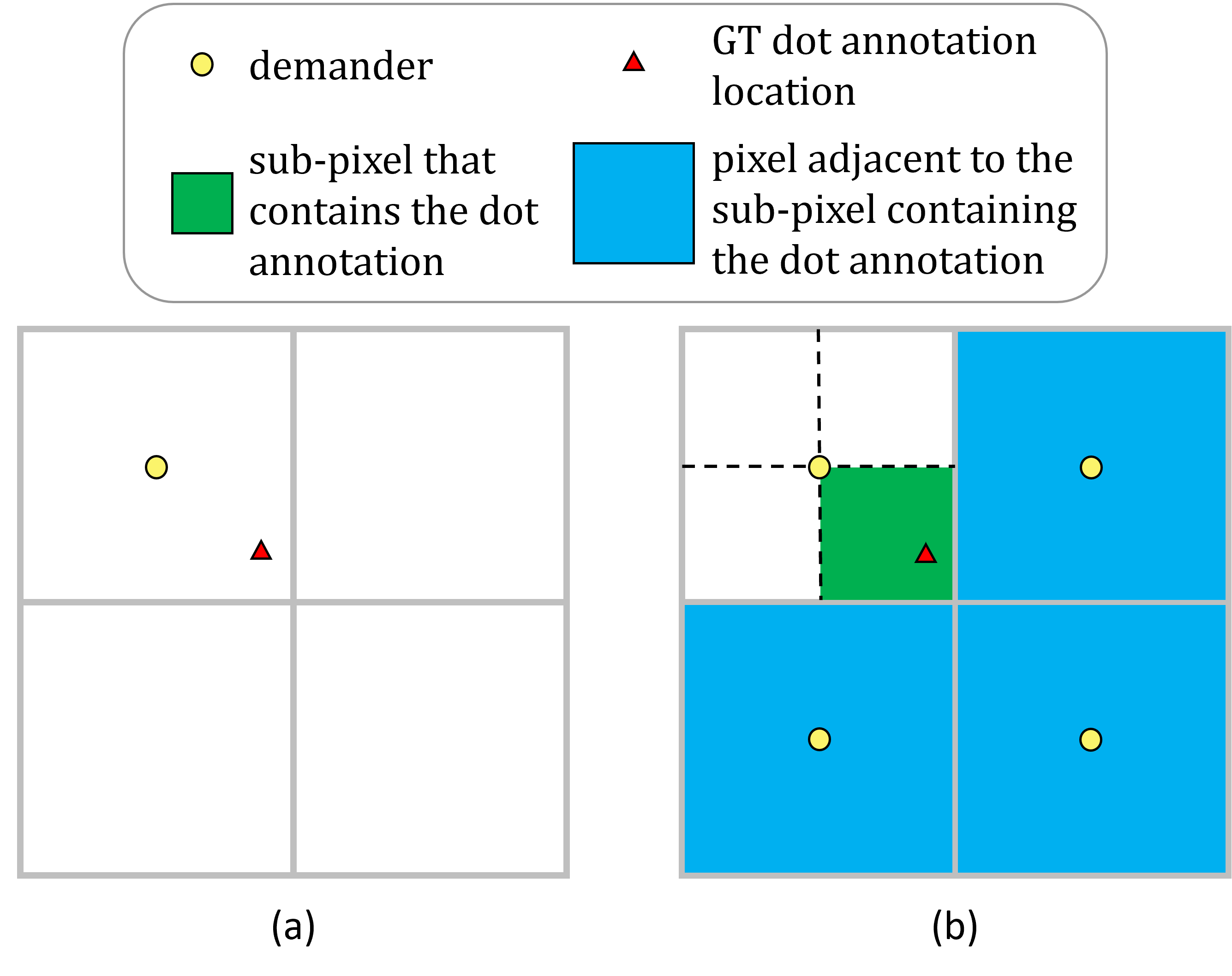}
\caption{Illustration of (a) the naive formulation of the demanders $D$, and (b) formulation of the demanders $D$ that involves the idea of sub-pixel resolution.}
\label{fig:demander}
\end{wrapfigure}

\noindent\textbf{The demanders $D$.} As for the demanders, we aim to construct $D$ w.r.t. each body joint to represent its corresponding GT dot annotation.
To achieve this goal, for each body joint, a naive formulation of $D$ is to identify the pixel containing the GT dot annotation (i.e., the upper left pixel in Fig.~\ref{fig:demander}(a)) and construct a single demander (i.e., the yellow dot in Fig.~\ref{fig:demander}(a)) at the center of this pixel.
However, this naive formulation can result in a suboptimal model performance, as the demander formulated in this way can be noticeably different from what the GT dot annotation might suggest.
Generally, the predicted heatmaps outputted by most of the existing \textit{heatmap-based methods} \cite{tompson2014joint,newell2016stacked,xiao2018simple,sun2019deep,li2019rethinking,cheng2020higherhrnet,zhang2020distribution,yang2020transpose,li2021tokenpose,luo2021rethinking,YuanFHLZCW21} have a lower resolution compared to the input image. For example, for the method HRNet \cite{sun2019deep}, when the size of the input image is $384\times288$, the size of the predicted heatmap is $96\times72$ only. Due to such a resolution gap, as shown in  Fig.~\ref{fig:demander}(a), there can exist a non-negligible distance between the location of the demander and the location of the dot annotation, which can affect the performance of the pose estimation model.

To address this problem, we aim to make the formulated demanders $D$ a more accurate representation of the dot annotation by overcoming the resolution gap. To achieve this, inspired by the fact that the more accurate location information of an object can be deduced using sub-pixel resolutions \cite{khademi2012sub}, we propose to formulate the demanders $D$ with the following four steps. (1) As illustrated in the upper left of Fig.~\ref{fig:demander}(b), we first split the pixel containing the dot annotation into four "sub-pixels" (four squares separated from each other by dashed lines). (2) After that, we identify the "sub-pixel" that the dot annotation lies in (i.e., the green "sub-pixel" shown in Fig.~\ref{fig:demander}(b)). (3) Then, as shown by the four yellow dots in Fig.~\ref{fig:demander}(b), we localize a set of four demanders at the centers of both the pixel containing the identified "sub-pixel" (i.e., the upper left pixel in Fig.~\ref{fig:demander}(b)), and the three pixels adjacent to the identified "sub-pixel" (i.e., the three blue pixels in Fig.~\ref{fig:demander}(b)). (4) Finally, 
we determine the units of mass each demander requires in the following way so that the dot annotation can be precisely represented using the set of 
the four demanders.
Specifically, 
denote the 2D Euclidean distance between the centers of two adjacent pixels as $g$ (which is also the pixel size), the coordinates of the GT dot annotation as $(x_{dot}, y_{dot})$, and the coordinates of the $i$-th demander as $(x_i, y_i)$, where $i \in \{1, 2, 3, 4\}$. Then we construct the demanders $D$ as:
\begin{equation}\label{eq:demander}
\begin{aligned}
D = (d_1, d_2, d_3, d_4) ,~\textbf{where}~ d_i = \frac{(g-|x_{dot} - x_i|) \times (g-|y_{dot} - y_i|)}{g^2}
\end{aligned}
\end{equation}
where $d_i$ denotes the units of mass the $i$-th demander requires. 
By designing $d_i$ in this way, we can achieve the following two properties. (1) Among the four demanders, via assigning more mass to the demanders nearer the dot annotation and less mass to the demanders farther from the dot annotation, we can accurately derive the coordinates of the dot annotation as $(x_{dot}, y_{dot}) = (d_1 \times x_1 + d_2 \times x_2 + d_3 \times x_3 + d_4 \times x_4, d_1 \times y_1 + d_2 \times y_2 + d_3 \times y_3 + d_4 \times y_4)$. Hence, the set of demanders $D$ constructed in this way can represent the GT dot annotation accurately, regardless of how many times the resolution of the heatmap is lower than the resolution of the input image. (2) Besides, as $d_1 + d_2 + d_3 + d_4 = 1$, the total units of mass required by the demanders $D$ are also constrained to be the same as the total units of mass stored by the suppliers $S$.

\noindent\textbf{The cost function $C$.} 
To measure the distribution difference between the $H \times W$ suppliers ($S$) constructed from the predicted heatmap and the $4$ demanders ($D$) constructed from the GT dot annotation via calculating their Earth Mover's Distance, we also need to formulate a cost function $C \in  R^{HW \times 4}_{\geq 0}$. 
Denote the coordinates of the $n$-th supplier as $(x_{s_n}, y_{s_n})$, where $n \in \{1, ..., H \times W\}$, and the coordinates of the $m$-th demander as $(x_{d_m}, y_{d_m})$, where $m \in \{1, 2, 3, 4\}$. We here simply formulate $C_{n,m}$, the cost per unit transported from the $n$-th supplier to the $m$-th demander, as the L2 distance between the $n$-th supplier and the $m$-th demander, i.e., $C_{n,m} = \sqrt{(x_{d_m} - x_{s_n})^2 + (y_{d_m} - y_{s_n})^2}$. Using such a cost function, our method can optimize the model directly towards accurately localizing the dot annotation of the body joint via minimizing the distribution difference between the suppliers $S$ and the demanders $D$.

\subsection{Loss Function}
\label{Sec:loss}

Above we formulate the optimization of the heatmap prediction as a distribution matching problem for a single body joint. In this section, we introduce how we construct the loss function following such a formulation. Note that in 2D human pose estimation, we need to locate multiple body joints. Therefore, to construct the loss function, we first construct the suppliers, the demanders, and the cost function for each body joint respectively. After that, we calculate the loss value corresponding to each body joint via measuring the distribution difference (calculating the Earth Mover's Distance) between its corresponding demanders and suppliers.
Specifically, denote $K$ the total number of body joints and $L_k$ the loss term calculated w.r.t. the $k$-th body joint. We then formulate the loss function as:  
\begin{equation}\label{eq:loss}
\begin{aligned}
L_{Matching} = \sum_{k=1}^{K} L_k,~\textbf{where}~L_k~=~E^{reg}_{C^k}(S^k, D^k)
\end{aligned}
\end{equation}
where $S^k$, $D^k$, and $C^k$ respectively denote the constructed suppliers, demanders, and cost function for the $k$-th body joint. The corresponding Earth Mover's Distance $E^{reg}_{C^k}(S^k, D^k)$ is calculated using Eq.~\ref{eq:pre_2}.

\subsection{Training and Testing}
\label{Sec:whole}

Our proposed method can be flexibly applied on various off-the-shelf 2D human pose estimation models. During training, we optimize the heatmap prediction via minimizing the loss function in Eq.~\ref{eq:loss}.
During testing, for each body joint, we first select a square of four adjacent pixels with the largest sum of pixel values from the predicted heatmap, and then normalize the sum of these four pixel values to 1. The normalized pixel values of these four pixels can be seen as the units of mass each of them requires. Hence, in the same way as how we get the coordinates of the GT dot annotation from the demanders in Sec.~\ref{Sec:form}, we can get the predicted coordinates of this joint. Specifically, denote the coordinates of the $i$-th selected pixel as $(x_i, y_i)$, and the normalized pixel value of this pixel as $d_i$, where $i \in \{1, 2, 3, 4\}$. We can derive the predicted coordinates of this joint as $(x_{pred}, y_{pred}) = (d_1 \times x_1 + d_2 \times x_2 + d_3 \times x_3 + d_4 \times x_4, d_1 \times y_1 + d_2 \times y_2 + d_3 \times y_3 + d_4 \times y_4)$.

\section{Analysis}
\label{Sec:proof}

Most of the \textit{heatmap-based methods} \cite{tompson2014joint,newell2016stacked,xiao2018simple,sun2019deep,li2019rethinking,cheng2020higherhrnet,zhang2020distribution,yang2020transpose,li2021tokenpose,luo2021rethinking,YuanFHLZCW21} optimize the heatmap prediction via minimizing the pixel-wise MSE loss between the predicted heatmap and the Gaussian-smoothed heatmap. Below we do some analysis about this type of methods. 

We denote $K$ the number of joints per input image $I$. Then we denote $\mathbf{H_{dot}} = \{H_{dot}^1, ..., H_{dot}^K\}$, $\mathbf{H_{Gau}} = \{H_{Gau}^1, ..., H_{Gau}^K\}$,  and $\mathbf{H_{pred}} = \{H_{pred}^1, ..., H_{pred}^K\}$ respectively the corresponding $K$ GT dot-annotated heatmaps, GT Gaussian-smoothed heatmaps, and predicted heatmaps of the input image $I$. We denote $\mathcal D_{dot} = \{(I, \mathbf{H_{dot}})\}$ the joint distribution of the input image and the corresponding $K$ dot-annotated heatmaps, and $\mathcal D_{Gau} = \{(I, \mathbf{H_{Gau}})\}$ the joint distribution of the input image and the corresponding $K$ Gaussian-smoothed heatmaps. Besides, we denote $l_{MSE}(a, b) = \left\| a - b\right\|_2^2$ the pixel-wise MSE loss, and $\phi$ the model parameters where $\phi(I) = \mathbf{H_{pred}}$. After that, we denote $R(\mathcal D_{dot}, \phi, l_{MSE}) = \mathbb{E}_{(I, \mathbf{H_{dot}}) \sim \mathcal D_{dot}}[l_{MSE}(\phi(I), \mathbf{H_{dot}})]$ as the expected risk calculated between the predicted heatmaps and the GT dot-annotated heatmaps, and $R(\mathcal D_{Gau}, \phi, l_{MSE}) = \mathbb{E}_{(I, \mathbf{H_{Gau}}) \sim \mathcal D_{Gau}}[l_{MSE}(\phi(I), \mathbf{H_{Gau}})]$ as the expected risk calculated between the predicted heatmaps and the Gaussian-smoothed heatmaps.

\begin{theorem}
\label{thm:theorem_1}
The relationship between $R(\mathcal D_{dot}, \phi, l_{MSE})$ and $R(\mathcal D_{Gau}, \phi, l_{MSE})$ can be written as
\begin{equation}\label{eq:theorem_1}
\begin{aligned}
R(\mathcal D_{dot}, \phi, l_{MSE}) = &R(\mathcal D_{Gau}, \phi, l_{MSE}) + 2 \times \mathbb{E}_{(I, \mathbf{H_{Gau}}) \sim \mathcal D_{Gau}}[\langle\mathbf{H_{pred}},\mathbf{H_{Gau}}\rangle] \\& - 2 \times \mathbb{E}_{(I, \mathbf{H_{dot}}) \sim \mathcal D_{dot}} [\langle\mathbf{H_{pred}},\mathbf{H_{dot}}\rangle] - C
\end{aligned}
\end{equation}
where 
$C = \mathbb{E}_{(I, \mathbf{H_{dot}}) \sim \mathcal D_{dot}}[\left\|\mathbf{H_{Gau}} - \mathbf{H_{dot}}\right\|_2^2]$ is a constant.
\end{theorem}

The proof of Theorem~\ref{thm:theorem_1} is provided in the supplementary.
As shown in Theorem~\ref{thm:theorem_1}, when we minimize the pixel-wise loss between $\mathbf{H_{pred}}$ and $\mathbf{H_{Gau}}$, while $R(\mathcal D_{Gau}, \phi, l_{MSE})$ decreases, the second term $2 \times \mathbb{E}_{(I, \mathbf{H_{Gau}}) \sim \mathcal D_{Gau}}[\langle\mathbf{H_{pred}},\mathbf{H_{Gau}}\rangle]$ can increase and the third term $2 \times \mathbb{E}_{(I, \mathbf{H_{dot}}) \sim \mathcal D_{dot}} [\langle\mathbf{H_{pred}},\mathbf{H_{dot}}\rangle]$ cannot be guaranteed to increase or decrease. Because of this, $R(\mathcal D_{dot}, \phi, l_{MSE})$ cannot be guaranteed to decrease, and the model performance of body joint localization may not be consistently improved during such optimization of heatmap prediction. A more intuitive analysis is as follows. In the optimization process of most of the \textit{heatmap-based methods}, since all the $H \times W$ pixels from the predicted heatmap contribute to the overall pixel-wise loss, learning to fit the other pixels better instead of the pixel representing the dot annotation can also lead to a smaller overall loss, as shown in Fig.~\ref{fig:example}. Hence, when minimizing the overall pixel-wise loss, the model performance of body joint localization cannot be guaranteed to improve consistently. Besides, during training, the GT Gaussian-smoothed heatmap is constructed via using the Gaussian blob. Therefore, during testing, the predicted heatmap often has a relatively large area of pixels with large values around the dot annotation, as shown in Fig.~\ref{fig:visual}, which can make it difficult to accurately locate the body joint.

Differently, our method optimizes the heatmap prediction from a novel perspective via minimizing the difference between the distribution constructed from the predicted heatmap (i.e., the suppliers $S$), and the distribution constructed from the dot annotation (i.e., the demanders $D$). During training, 
since we formulate the cost function between each pair of supplier and demander as their L2 distance, minimizing the distribution difference based on such a cost function can aggregate the pixel values in the predicted heatmap towards the dot annotation. Hence, our method can help to achieve a more consistent model performance improvement when minimizing the distribution difference. Besides, during testing, equipped with such a mechanism of aggregating the predicted pixel values towards the dot annotation,
our method can achieve a more compact body joint localization as shown in Fig.~\ref{fig:visual}.

\begin{table}
\caption{The improvement of AP on COCO validation set when our proposed method is applied to various baselines.}
\footnotesize
\centering
\resizebox{\linewidth}{!}{
\begin{tabular}{l|l|l|c|lccccc}
\hline
Method & Venue & Backbone &Input size & AP & $\text{AP}^{50}$ & $\text{AP}^{75}$ & $\text{AP}^{\text{M}}$ &$\text{AP}^{\text{L}}$ &AR  \\
\hline
Hourglass\cite{newell2016stacked}& ECCV~2016 & 8-Stage Hourglass &$256\times192$ &66.9 & - & - & - & - & -\\
CPN\cite{chen2018cascaded}& CVPR~2018 & ResNet-50 &$256\times192$ &69.4 & - & - & - & - & -\\
CPN\cite{chen2018cascaded}& CVPR~2018 & ResNet-50 &$384\times288$ &71.6 & - & - & - & - & -\\
DarkPose\cite{zhang2020distribution}& CVPR~2020 & HRNet-W32   &$384\times288$ &76.6&90.7&82.8&72.7&83.9&81.5\\
UDP\cite{huang2020devil}& CVPR~2020 & HRNet-W32 &$384\times288$ &77.8&91.7&84.5&74.2&84.3&82.4\\
UDP\cite{huang2020devil}& CVPR~2020 & HRNet-W48 &$384\times288$ &77.8&92.0&84.3&74.2&84.5&82.5\\
TokenPose\cite{li2021tokenpose}& ICCV~2021 & TokenPose-L/D24 & $256\times192$ & 75.8 &90.3 &82.5 &72.3 &82.7 &80.9\\
Removing Bias\cite{gu2021removing}& ICCV~2021 & ResNet-152   &$384\times288$ & 74.4 & - & - & - & - & -\\
Removing Bias\cite{gu2021removing}& ICCV~2021 & HRNet-W32   &$256\times192$ & 75.8 & - & - & - & - & -\\
\hline
Simple Baseline\cite{xiao2018simple} & ECCV~2018 & ResNet-152   &$384\times288$ & 75.0 & 90.8 & 82.1 & 67.8 & 78.3 & 80.0 \\
\textbf{+ Ours} & & ResNet-152   &$384\times288$ & \textbf{76.7(\textuparrow 1.7)} & \textbf{92.1} & \textbf{83.6} & \textbf{69.7} & \textbf{80.0} & \textbf{81.3}\\
\hline
HRNet\cite{sun2019deep} & CVPR~2019 & HRNet-W32 &$384\times288$ &76.7&91.9&83.6&73.2&83.2&81.6 \\
\textbf{+ Ours} & & HRNet-W32 &$384\times288$ & \textbf{78.2(\textuparrow1.5)} & \textbf{92.2} & \textbf{84.5} & \textbf{74.3} & \textbf{84.7} & \textbf{82.5}\\
\hline
HRNet\cite{sun2019deep} & CVPR~2019 & HRNet-W48 &$384\times288$ &77.1&91.8&83.8&73.5&83.5&81.8\\
\textbf{+ Ours} & & HRNet-W48 &$384\times288$ & \textbf{78.8(\textuparrow1.7)} & \textbf{92.5} & \textbf{85.1} & \textbf{75.0} & \textbf{85.3} & \textbf{83.1}\\
\hline
HRFormer\cite{YuanFHLZCW21} & NIPS~2021 & HRFormer-Base &$384\times288$ &78.0&92.2&84.8& 74.3 & 84.6 &82.6\\
\textbf{+ Ours} & & HRFormer-Base &$384\times288$ & \textbf{78.9(\textuparrow0.9)} & \textbf{92.6} & \textbf{85.4} & \textbf{75.3} & \textbf{85.3} & \textbf{83.3}\\
\hline
\end{tabular}}
\label{Tab:coco_val}
\end{table}

\section{Experiments}

To evaluate the effectiveness of our proposed method, we conduct experiments on the COCO dataset \cite{lin2014microsoft} and the MPII Human Pose dataset \cite{andriluka20142d}. Besides, to test the generality of our method, we apply it to various backbones, e.g., ResNet \cite{xiao2018simple}, HRNet \cite{sun2019deep}, and HRFormer \cite{YuanFHLZCW21}.

\subsection{COCO Keypoint Detection}

\noindent\textbf{Dataset \& evaluation metric.} The COCO dataset \cite{lin2014microsoft} contains more than 200k images and 250k person instances, which are annotated with 17 body joints. This dataset has three subsets including COCO training set, COCO validation set, and COCO test-dev set, which have 57k, 5k and 20k images, respectively. We conduct experiments on this dataset via first training the model on the train2017 set, and then evaluating the model on the val2017 set and test-dev2017 set. Following \cite{xiao2018simple,sun2019deep,YuanFHLZCW21}, we use standard average precision (AP) calculated based on Object Keypoint Similarity (OKS) to evaluate model performance.

\noindent\textbf{Implementation details.} We apply our method to various baselines including Simple Baseline \cite{xiao2018simple}, HRNet \cite{sun2019deep}, and HRFormer \cite{YuanFHLZCW21}, with their respective backbones including ResNet-152, HRNet-W32, HRNet-W48, and HRFormer-Base. For these baselines, we follow their original learning and optimization configurations for model training. 
To calculate the Earth Mover's Distance using the Sinkhorn algorithm, we set the Sinkhorn entropic regularization parameter to 1 and the number of Sinkhorn iterations to 1000 in our experiments.

\noindent\textbf{Results.} In Tab.~\ref{Tab:coco_val} and Tab.~\ref{Tab:coco_test_dev}, we report results on the COCO validation and test-dev sets. We observe that after applying our method on various baselines, a significant performance enhancement is achieved, which shows effectiveness of our proposed method. Moreover, we compare our method with other state-of-the-art 2D human pose estimation methods. Our method achieves superior performance compared to these methods, further demonstrating the effectiveness of our  method.

\begin{table}
\caption{The improvement of AP on COCO test-dev set when our proposed method is applied to various baselines.}
\footnotesize
\centering
\resizebox{\linewidth}{!}{
\begin{tabular}{l|l|l|c|lccccc}
\hline
Method & Venue & Backbone &Input size & AP & $\text{AP}^{50}$ & $\text{AP}^{75}$ & $\text{AP}^{\text{M}}$ &$\text{AP}^{\text{L}}$ &AR  \\
\hline
G-RMI\cite{papandreou2017towards} & CVPR~2017 & ResNet-101 &$353\times 257$ &64.9 & 85.5 & 71.3   & 62.3 &70.0 &69.7\\
Mask-RCNN\cite{he2017mask} & ICCV~2017  & ResNet-50-FPN  &- &63.1 & 87.3 & 68.7 & 57.8 &71.4 &- \\
RMPE\cite{fang2017rmpe}  & ICCV~2017 & PyraNet\cite{yang2017learning}&$320\times 256$  &72.3 & 89.2   & 79.1   & 68.0 &78.6 &-   \\
CFN\cite{huang2017coarse} & ICCV~2017  & -      &-  &72.6 & 86.1   & 69.7   & 78.3 &64.1 &-   \\
CPN\cite{chen2018cascaded} & CVPR~2018 & ResNet-Inception &$384\times288$ &72.1 & 91.4   & 80.0   & 68.7 &77.2 &78.5\\
CPN(ensemble)\cite{chen2018cascaded} & CVPR~2018 & ResNet-Inception &$384\times 288$       &73.0 & 91.7   & 80.9   & 69.5 &78.1 &79.0\\
Integral Pose Regression\cite{sun2018integral} & ECCV~2018  & ResNet-101 &$256\times 256$ &67.8 & 88.2 & 74.8 & 63.9 &74.0 &-   \\
Posefix\cite{moon2019posefix}  & CVPR~2019  & ResNet-152       &$384\times288$      &73.6 & 90.8   & 81.0   & 70.3 &79.8 &79.0\\
DarkPose\cite{zhang2020distribution}& CVPR~2020& HRNet-W48 &$384\times288$& 76.2 & 92.5 & 83.6 & 72.5 & 82.4 & 81.1\\
UDP\cite{huang2020devil} & CVPR~2020 & HRNet-W48 &$384\times288$ &76.5   & 92.7 & 84.0 & 73.0&82.4 &81.6\\
TokenPose\cite{li2021tokenpose}& ICCV~2021 & TokenPose-L/D24 & $384\times288$ & 75.9 & 92.3 & 83.4 & 72.2 & 82.1 & 80.8\\
Removing Bias\cite{gu2021removing}& ICCV~2021 & HRNet-W48 &$384\times288$ & 76.1 & - & - & - & - & 81.0\\
\hline
Simple Baseline\cite{xiao2018simple}& ECCV~2018 & ResNet-152   &$384\times288$ &73.8 & 91.7 & 81.2 & 70.3 &80.0 &79.1\\
\textbf{+ Ours} && ResNet-152   &$384\times288$ & \textbf{75.3(\textuparrow1.5)} & \textbf{92.6} & \textbf{83.1} & \textbf{71.7} & \textbf{81.1} & \textbf{80.3}\\
\hline
HRNet\cite{sun2019deep} & CVPR~2019 & HRNet-W32 &$384\times288$ &74.9 & 92.5 & 82.8 & 71.3 &80.9 &80.1\\
\textbf{+ Ours} && HRNet-W32 &$384\times288$ & \textbf{76.7(\textuparrow1.8)} & \textbf{92.6} & \textbf{84.0} & \textbf{73.0} & \textbf{82.8} & \textbf{81.5}\\
\hline
HRNet\cite{sun2019deep} & CVPR~2019& HRNet-W48 &$384\times288$ &75.5 & 92.5  & 83.3& 71.9  &81.5 &80.5\\
\textbf{+ Ours} && HRNet-W48 &$384\times288$ & \textbf{77.2(\textuparrow1.7)} & \textbf{93.0} & \textbf{84.4} & \textbf{73.4} & \textbf{83.3} & \textbf{82.0}\\
\hline
HRFormer\cite{YuanFHLZCW21} & NIPS~2021& HRFormer-Base &$384\times288$ & 76.2 & 92.7 & 83.8 & 72.5 & 82.3 & 81.2\\
\textbf{+ Ours} & & HRFormer-Base &$384\times288$ & \textbf{77.2(\textuparrow1.0)} & \textbf{93.1} & \textbf{84.7} & \textbf{73.8} & \textbf{83.0} & \textbf{82.1}\\
\hline
\end{tabular}}
\label{Tab:coco_test_dev}
\end{table}

\subsection{MPII Human Pose Estimation}

\noindent\textbf{Dataset \& evaluation metric.} The MPII dataset \cite{andriluka20142d} contains around 25K images and more than 40k person instances, which are annotated with 16 body joints. We adopt the standard train/val split in \cite{andriluka20142d} to build the MPII training set and validation set, and conduct all the experiments on this dataset via first training the model on the MPII training set, and then evaluating it on the MPII validation set. Following \cite{sun2019deep}, we use the head-normalized probability of correct keypoint (PCKh) \cite{andriluka20142d} score as the evaluation metric on this dataset and report the PCKh@0.5 score.

\noindent\textbf{Implementation details.} On the MPII dataset, we also apply our method to various methods as our baselines, including Simple Baseline \cite{xiao2018simple} and HRNet \cite{sun2019deep}, with their respective backbones including ResNet-152, HRNet-W32, and HRNet-W48. We follow the original learning and optimization configurations for model training for both Simple Baseline \cite{xiao2018simple} and HRNet \cite{sun2019deep}. Besides, same as the experiments on the COCO dataset, we also set the Sinkhorn entropic regularization parameter to 1 and the number of Sinkhorn iterations to 1000 in our experiments on the MPII dataset.

\noindent\textbf{Results on the MPII validation set.} In Tab.~\ref{Tab:mpii}, we report the results on the MPII validation set. As shown, applying our proposed method on various baselines results in a consistent performance improvement, which demonstrates the effectiveness of our proposed method.

\begin{table}[t]
\caption{The improvement of AP on MPII validation set when our proposed method is applied to various baselines.}
\footnotesize
\centering
\resizebox{\linewidth}{!}{
\begin{tabular}{l|l|l|c|lccccccc}
\hline
Method & Venue & Backbone &Input size & Mean & Hea & Sho & Elb & Wri & Hip & Kne & Ank \\
\hline
Integral Pose Regression\cite{sun2018integral} & ECCV~2018  & ResNet-101 &$256\times 256$ &87.9&-&-&-&-&-&-&-\\
UDP\cite{huang2020devil} & CVPR~2020& HRNet-W32 &$256\times256$ &90.4&97.4&96.0&91.0&86.5&89.1&86.6&83.3\\
DarkPose\cite{zhang2020distribution} & CVPR~2020 & HRNet-W32   &$256\times256$ &90.6&97.2&95.9&91.2&86.7&89.7&86.7&84.0\\
TokenPose\cite{li2021tokenpose} & ICCV~2021& TokenPose-L/D6 & $256\times256$  &90.1&97.1&95.9&91.0&85.8&89.5&86.1&82.7 \\
TokenPose\cite{li2021tokenpose} & ICCV~2021& TokenPose-L/D12 & $256\times256$  &90.1&97.2&95.8&90.7&85.9&89.2&86.2&82.3 \\
TokenPose\cite{li2021tokenpose} & ICCV~2021& TokenPose-L/D24 & $256\times256$  &90.2&97.1&95.9&90.4&86.0&89.3&87.1&82.5 \\
Removing Bias\cite{gu2021removing} & ICCV~2021& ResNet-152   &$256\times256$ & 89.9 & - & - & - & - & -& - & -\\
Removing Bias\cite{gu2021removing} & ICCV~2021& HRNet-W32   &$256\times256$ & 90.6 & - & - & - & - & -& - & -\\
\hline
Simple Baseline\cite{xiao2018simple}& ECCV~2018 & ResNet-152   &$256\times256$ & 89.6 & \textbf{97.0} & 95.9 & 90.0 & 85.0 & \textbf{89.2} & 85.3 & 81.3  \\
\textbf{+ Ours} && ResNet-152   &$256\times256$ & \textbf{90.3(\textuparrow0.7)} & \textbf{97.0} & \textbf{96.1} & \textbf{90.6} & \textbf{86.1} & \textbf{89.2} & \textbf{86.7} & \textbf{83.1} \\
\hline
HRNet\cite{sun2019deep} & CVPR~2019& HRNet-W32 &$256\times256$ & 90.4 & 97.1 & 95.9 & 90.7 & 86.1 & 89.4 & 86.9 & 83.2 \\
\textbf{+ Ours}& & HRNet-W32 &$256\times256$ & \textbf{90.9(\textuparrow0.5)} & \textbf{97.3} & \textbf{96.2} & \textbf{91.2} & \textbf{86.8} & \textbf{90.1} & \textbf{87.4} & \textbf{84.1} \\
\hline
HRNet\cite{sun2019deep} & CVPR~2019& HRNet-W48 &$256\times256$ & 90.5 & 96.9 & 96.0 & 90.9 & 86.2 & 89.6 & 87.1 & 83.5 \\
\textbf{+ Ours} && HRNet-W48 &$256\times256$ & \textbf{90.9(\textuparrow0.4)} & \textbf{97.1} & \textbf{96.3} & \textbf{91.2} & \textbf{87.0} & \textbf{90.2} & \textbf{87.5} & \textbf{84.2}\\
\hline
\end{tabular}}
\label{Tab:mpii}
\end{table}

\subsection{Ablation Studies}
We conduct ablation studies on the COCO validation set via applying our proposed method on HRNet-W48 \cite{sun2019deep}.

\noindent\textbf{Impact of involving the idea of sub-pixels in formulating $D$.} In our proposed method, we formulate the demanders $D$ by involving the idea of sub-pixel resolution. To investigate the impact of formulating the demanders $D$ in such a way, we compare our proposed method (\textbf{sub-pixel demanders formulation}) with a variant (\textbf{naive demanders formulation}). This variant still formulates the suppliers $S$ and the cost function $C$ in the same way, but formulates the demanders $D$ naively as a single demander at the center of the pixel containing the dot annotation, as shown in Fig.~\ref{fig:demander}(a). As shown in Tab.~\ref{Tab:ablation_1}, our proposed method consistently outperforms this variant, which shows effectiveness of our \textbf{sub-pixel demanders formulation}.

\begin{table}[t]
\parbox{0.49\textwidth}{
\caption{Evaluation on the effectiveness of formulating the demanders $D$ involving the idea of sub-pixel resolution.}
\label{Tab:ablation_1}
\resizebox{0.49\textwidth}{!}{
\begin{tabular}{l|cccccc}
\hline
Method & AP & $\text{AP}^{50}$ & $\text{AP}^{75}$ & $\text{AP}^{\text{M}}$ &$\text{AP}^{\text{L}}$ &AR  \\
\hline
Baseline(HRNet-W48) &77.1 &91.8 &83.8 &73.5 &83.5 &81.8\\
\hline
\textbf{Naive demanders formulation} & 77.9 & 92.5 & 84.8 & 74.5 & 83.9 & 82.4 \\
\textbf{Sub-pixel demanders formulation} & 78.8 & 92.5 & 85.1 & 75.0 & 85.3 & 83.1\\
\hline
\end{tabular}}}
\hspace{0.01\textwidth}
\parbox{0.49\textwidth}{
\caption{Evaluation on the number of Sinkhorn iterations.}
\label{Tab:ablation_2}
\resizebox{0.49\textwidth}{!}{
\begin{tabular}{l|cccccc}
\hline
Method & AP & $\text{AP}^{50}$ & $\text{AP}^{75}$ & $\text{AP}^{\text{M}}$ &$\text{AP}^{\text{L}}$ &AR  \\
\hline
Baseline(HRNet-W48) &77.1 &91.8 &83.8 &73.5 &83.5 &81.8\\
\hline
500 Sinkhorn iterations & 78.3 & 92.4 & 84.9 & 74.8 & 84.4 & 82.7 \\
1000 Sinkhorn iterations & 78.8 & 92.5 & 85.1 & 75.0 & 85.3 & 83.1\\
1500 Sinkhorn iterations & 78.7 & 92.4 & 85.2 & 74.8 & 85.4 & 83.0 \\
\hline
\end{tabular}}}
\end{table}

\begin{wrapfigure}[19]{r}{0.54\textwidth}
\vspace{-0.15cm}
\centering
\includegraphics[width=0.54\textwidth]{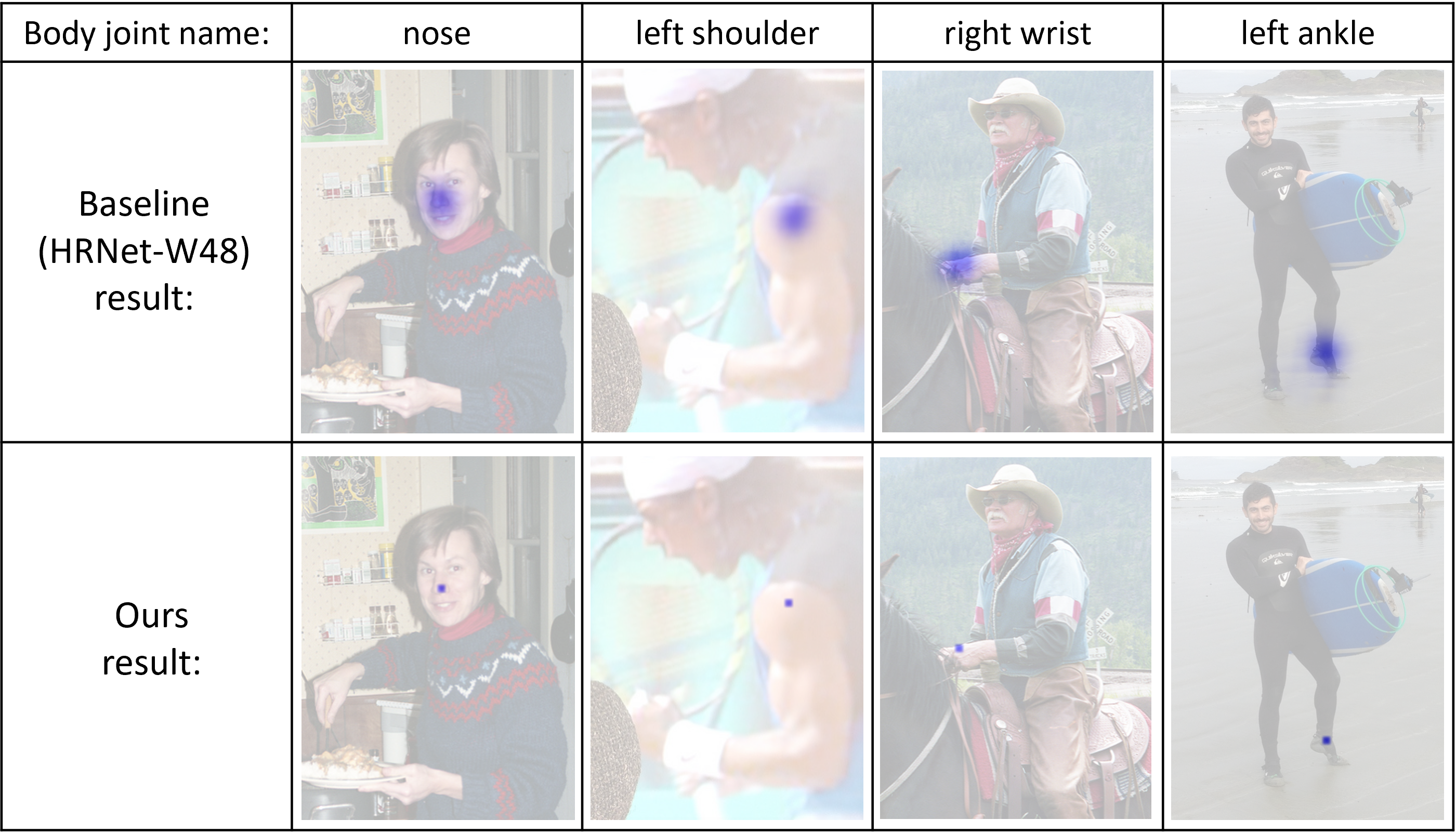}
\caption{Qualitative results of our method formulating the optimization of the heatmap prediction as a distribution matching problem and the baseline method \cite{sun2019deep} using the Gaussian-smoothed heatmap as the optimization objective. As shown, our method localizes body joints much more compactly.}
\label{fig:visual}
\end{wrapfigure}

\noindent\textbf{Impact of the number of Sinkhorn iterations.} For measuring the Earth Mover's Distance utilizing the Sinkhorn algorithm, we need to set the number of Sinkhorn iterations, which we set to 1000 in our experiments. We evaluate other choices of the number of Sinkhorn iterations in Tab.~\ref{Tab:ablation_2}. As shown, all variants outperform the baseline method, and after the number of Sinkhorn iterations becomes larger than 1000, the model performance becomes stabilized. Hence, we set the number of Sinkhorn iterations to be 1000 in all our experiments.

\noindent\textbf{Qualitative results.} 
Some qualitative results are shown in Fig.~\ref{fig:visual}. As shown, via formulating the cost function as the L2 distance, our proposed method can aggregate the pixel values in the predicted heatmap towards the dot annotation, and thus localize body joints much more compactly than the baseline method \cite{sun2019deep} relying on the Gaussian-smoothed heatmap. This demonstrates that our method can effectively optimize the model in the direction of accurately localizing the body joint.

\section{Conclusion}
In this paper, from a novel perspective, we formulate the optimization of the heatmap prediction as a distribution matching problem between the predicted heatmap and the dot annotation via calculating their Earth Mover's Distance. Our proposed method is simple yet effective, and can be easily applied to various 2D human pose estimation models. Our method achieves superior performance on the COCO dataset and the MPII dataset.

\begin{ack}
This project is supported by the Ministry of Education, Singapore, under the SUTD Kickstarter Initiative Project (SKI 2021\_02\_06), National Research Foundation Singapore under its AI Singapore Programme (AISG-100E-2020-065), and SUTD Startup Research Grant.
\end{ack}

\bibliographystyle{plain}
\bibliography{egbib}

\begin{thebibliography}{10}

\bibitem{andriluka20142d}
Mykhaylo Andriluka, Leonid Pishchulin, Peter Gehler, and Bernt Schiele.
\newblock 2d human pose estimation: New benchmark and state of the art
  analysis.
\newblock In {\em Proceedings of the IEEE Conference on computer Vision and
  Pattern Recognition}, pages 3686--3693, 2014.

\bibitem{carreira2016human}
Joao Carreira, Pulkit Agrawal, Katerina Fragkiadaki, and Jitendra Malik.
\newblock Human pose estimation with iterative error feedback.
\newblock In {\em Proceedings of the IEEE conference on computer vision and
  pattern recognition}, pages 4733--4742, 2016.

\bibitem{chen2018cascaded}
Yilun Chen, Zhicheng Wang, Yuxiang Peng, Zhiqiang Zhang, Gang Yu, and Jian Sun.
\newblock Cascaded pyramid network for multi-person pose estimation.
\newblock In {\em Proceedings of the IEEE conference on computer vision and
  pattern recognition}, pages 7103--7112, 2018.

\bibitem{cheng2020higherhrnet}
Bowen Cheng, Bin Xiao, Jingdong Wang, Honghui Shi, Thomas~S Huang, and Lei
  Zhang.
\newblock Higherhrnet: Scale-aware representation learning for bottom-up human
  pose estimation.
\newblock In {\em Proceedings of the IEEE/CVF conference on computer vision and
  pattern recognition}, pages 5386--5395, 2020.

\bibitem{cuturi2013sinkhorn}
Marco Cuturi.
\newblock Sinkhorn distances: Lightspeed computation of optimal transport.
\newblock {\em Advances in neural information processing systems}, 26, 2013.

\bibitem{fang2017rmpe}
Hao-Shu Fang, Shuqin Xie, Yu-Wing Tai, and Cewu Lu.
\newblock Rmpe: Regional multi-person pose estimation.
\newblock In {\em Proceedings of the IEEE international conference on computer
  vision}, pages 2334--2343, 2017.

\bibitem{gu2021removing}
Kerui Gu, Linlin Yang, and Angela Yao.
\newblock Removing the bias of integral pose regression.
\newblock In {\em Proceedings of the IEEE/CVF International Conference on
  Computer Vision}, pages 11067--11076, 2021.

\bibitem{gu2022dive}
Kerui Gu, Linlin Yang, and Angela Yao.
\newblock Dive deeper into integral pose regression.
\newblock In {\em International Conference on Learning Representations}, 2022.

\bibitem{he2017mask}
Kaiming He, Georgia Gkioxari, Piotr Doll{\'a}r, and Ross Girshick.
\newblock Mask r-cnn.
\newblock In {\em Proceedings of the IEEE international conference on computer
  vision}, pages 2961--2969, 2017.

\bibitem{huang2020devil}
Junjie Huang, Zheng Zhu, Feng Guo, and Guan Huang.
\newblock The devil is in the details: Delving into unbiased data processing
  for human pose estimation.
\newblock In {\em Proceedings of the IEEE/CVF conference on computer vision and
  pattern recognition}, pages 5700--5709, 2020.

\bibitem{huang2017coarse}
Shaoli Huang, Mingming Gong, and Dacheng Tao.
\newblock A coarse-fine network for keypoint localization.
\newblock In {\em Proceedings of the IEEE international conference on computer
  vision}, pages 3028--3037, 2017.

\bibitem{khademi2012sub}
Siamak Khademi, Ahmad Darudi, and Zahra Abbasi.
\newblock A sub pixel resolution method.
\newblock {\em arXiv preprint arXiv:1211.2221}, 2012.

\bibitem{li2021human}
Jiefeng Li, Siyuan Bian, Ailing Zeng, Can Wang, Bo~Pang, Wentao Liu, and Cewu
  Lu.
\newblock Human pose regression with residual log-likelihood estimation.
\newblock In {\em Proceedings of the IEEE/CVF International Conference on
  Computer Vision}, pages 11025--11034, 2021.

\bibitem{li2019rethinking}
Wenbo Li, Zhicheng Wang, Binyi Yin, Qixiang Peng, Yuming Du, Tianzi Xiao, Gang
  Yu, Hongtao Lu, Yichen Wei, and Jian Sun.
\newblock Rethinking on multi-stage networks for human pose estimation.
\newblock {\em arXiv preprint arXiv:1901.00148}, 2019.

\bibitem{li2021tokenpose}
Yanjie Li, Shoukui Zhang, Zhicheng Wang, Sen Yang, Wankou Yang, Shu-Tao Xia,
  and Erjin Zhou.
\newblock Tokenpose: Learning keypoint tokens for human pose estimation.
\newblock In {\em Proceedings of the IEEE/CVF International Conference on
  Computer Vision}, pages 11313--11322, 2021.

\bibitem{lin2014microsoft}
Tsung-Yi Lin, Michael Maire, Serge Belongie, James Hays, Pietro Perona, Deva
  Ramanan, Piotr Doll{\'a}r, and C~Lawrence Zitnick.
\newblock Microsoft coco: Common objects in context.
\newblock In {\em European conference on computer vision}, pages 740--755.
  Springer, 2014.

\bibitem{luo2021rethinking}
Zhengxiong Luo, Zhicheng Wang, Yan Huang, Liang Wang, Tieniu Tan, and Erjin
  Zhou.
\newblock Rethinking the heatmap regression for bottom-up human pose
  estimation.
\newblock In {\em Proceedings of the IEEE/CVF Conference on Computer Vision and
  Pattern Recognition}, pages 13264--13273, 2021.

\bibitem{moon2019posefix}
Gyeongsik Moon, Ju~Yong Chang, and Kyoung~Mu Lee.
\newblock Posefix: Model-agnostic general human pose refinement network.
\newblock In {\em Proceedings of the IEEE/CVF Conference on Computer Vision and
  Pattern Recognition}, pages 7773--7781, 2019.

\bibitem{moryossef2020real}
Amit Moryossef, Ioannis Tsochantaridis, Roee Aharoni, Sarah Ebling, and Srini
  Narayanan.
\newblock Real-time sign language detection using human pose estimation.
\newblock In {\em European Conference on Computer Vision}, pages 237--248.
  Springer, 2020.

\bibitem{newell2016stacked}
Alejandro Newell, Kaiyu Yang, and Jia Deng.
\newblock Stacked hourglass networks for human pose estimation.
\newblock In {\em European conference on computer vision}, pages 483--499.
  Springer, 2016.

\bibitem{nibali2018numerical}
Aiden Nibali, Zhen He, Stuart Morgan, and Luke Prendergast.
\newblock Numerical coordinate regression with convolutional neural networks.
\newblock {\em arXiv preprint arXiv:1801.07372}, 2018.

\bibitem{papandreou2017towards}
George Papandreou, Tyler Zhu, Nori Kanazawa, Alexander Toshev, Jonathan
  Tompson, Chris Bregler, and Kevin Murphy.
\newblock Towards accurate multi-person pose estimation in the wild.
\newblock In {\em Proceedings of the IEEE conference on computer vision and
  pattern recognition}, pages 4903--4911, 2017.

\bibitem{peng2022optimal}
Hanyu Peng, Mingming Sun, and Ping Li.
\newblock Optimal transport for long-tailed recognition with learnable cost
  matrix.
\newblock In {\em International Conference on Learning Representations}, 2022.

\bibitem{rubner2000earth}
Yossi Rubner, Carlo Tomasi, and Leonidas~J Guibas.
\newblock The earth mover's distance as a metric for image retrieval.
\newblock {\em International journal of computer vision}, 40(2):99--121, 2000.

\bibitem{schulter2017deep}
Samuel Schulter, Paul Vernaza, Wongun Choi, and Manmohan Chandraker.
\newblock Deep network flow for multi-object tracking.
\newblock In {\em Proceedings of the IEEE Conference on Computer Vision and
  Pattern Recognition}, pages 6951--6960, 2017.

\bibitem{sun2019deep}
Ke~Sun, Bin Xiao, Dong Liu, and Jingdong Wang.
\newblock Deep high-resolution representation learning for human pose
  estimation.
\newblock In {\em Proceedings of the IEEE/CVF Conference on Computer Vision and
  Pattern Recognition}, pages 5693--5703, 2019.

\bibitem{sun2018integral}
Xiao Sun, Bin Xiao, Fangyin Wei, Shuang Liang, and Yichen Wei.
\newblock Integral human pose regression.
\newblock In {\em Proceedings of the European conference on computer vision
  (ECCV)}, pages 529--545, 2018.

\bibitem{szegedy2016rethinking}
Christian Szegedy, Vincent Vanhoucke, Sergey Ioffe, Jon Shlens, and Zbigniew
  Wojna.
\newblock Rethinking the inception architecture for computer vision.
\newblock In {\em Proceedings of the IEEE conference on computer vision and
  pattern recognition}, pages 2818--2826, 2016.

\bibitem{tompson2014joint}
Jonathan~J Tompson, Arjun Jain, Yann LeCun, and Christoph Bregler.
\newblock Joint training of a convolutional network and a graphical model for
  human pose estimation.
\newblock {\em Advances in neural information processing systems}, 27, 2014.

\bibitem{toshev2014deeppose}
Alexander Toshev and Christian Szegedy.
\newblock Deeppose: Human pose estimation via deep neural networks.
\newblock In {\em Proceedings of the IEEE conference on computer vision and
  pattern recognition}, pages 1653--1660, 2014.

\bibitem{wang2020distribution}
Boyu Wang, Huidong Liu, Dimitris Samaras, and Minh~Hoai Nguyen.
\newblock Distribution matching for crowd counting.
\newblock {\em Advances in Neural Information Processing Systems},
  33:1595--1607, 2020.

\bibitem{wei2020point}
Fangyun Wei, Xiao Sun, Hongyang Li, Jingdong Wang, and Stephen Lin.
\newblock Point-set anchors for object detection, instance segmentation and
  pose estimation.
\newblock In {\em European Conference on Computer Vision}, pages 527--544.
  Springer, 2020.

\bibitem{xiao2018simple}
Bin Xiao, Haiping Wu, and Yichen Wei.
\newblock Simple baselines for human pose estimation and tracking.
\newblock In {\em Proceedings of the European conference on computer vision
  (ECCV)}, pages 466--481, 2018.

\bibitem{yan2018spatial}
Sijie Yan, Yuanjun Xiong, and Dahua Lin.
\newblock Spatial temporal graph convolutional networks for skeleton-based
  action recognition.
\newblock In {\em Thirty-second AAAI conference on artificial intelligence},
  2018.

\bibitem{yang2020transpose}
Sen Yang, Zhibin Quan, Mu~Nie, and Wankou Yang.
\newblock Transpose: Towards explainable human pose estimation by transformer.
\newblock {\em arXiv e-prints}, pages arXiv--2012, 2020.

\bibitem{yang2017learning}
Wei Yang, Shuang Li, Wanli Ouyang, Hongsheng Li, and Xiaogang Wang.
\newblock Learning feature pyramids for human pose estimation.
\newblock In {\em proceedings of the IEEE international conference on computer
  vision}, pages 1281--1290, 2017.

\bibitem{YuanFHLZCW21}
Yuhui Yuan, Rao Fu, Lang Huang, Weihong Lin, Chao Zhang, Xilin Chen, and
  Jingdong Wang.
\newblock Hrformer: High-resolution transformer for dense prediction.
\newblock 2021.

\bibitem{zhang2020deepemd}
Chi Zhang, Yujun Cai, Guosheng Lin, and Chunhua Shen.
\newblock Deepemd: Few-shot image classification with differentiable earth
  mover's distance and structured classifiers.
\newblock In {\em Proceedings of the IEEE/CVF conference on computer vision and
  pattern recognition}, pages 12203--12213, 2020.

\bibitem{zhang2020distribution}
Feng Zhang, Xiatian Zhu, Hanbin Dai, Mao Ye, and Ce~Zhu.
\newblock Distribution-aware coordinate representation for human pose
  estimation.
\newblock In {\em Proceedings of the IEEE/CVF conference on computer vision and
  pattern recognition}, pages 7093--7102, 2020.

\bibitem{zhang2014pose}
Yizhai Zhang, Kuo Chen, Jingang Yi, and Liu Liu.
\newblock Pose estimation in physical human-machine interactions with
  application to bicycle riding.
\newblock In {\em 2014 IEEE/RSJ International Conference on Intelligent Robots
  and Systems}, pages 3333--3338. IEEE, 2014.

\bibitem{zhou2020occlusion}
Lu~Zhou, Yingying Chen, Yunze Gao, Jinqiao Wang, and Hanqing Lu.
\newblock Occlusion-aware siamese network for human pose estimation.
\newblock In {\em European Conference on Computer Vision}, pages 396--412.
  Springer, 2020.

\end{thebibliography}

\end{document}